\documentclass[letterpaper, 10pt, conference]{ieeeconf}
\IEEEoverridecommandlockouts
\usepackage{cite}
\usepackage{amsmath,amssymb,amsfonts}
\usepackage{algpseudocode}
\usepackage{algorithm}		%We use the package but force disable the float component for compliance with IEEETran
\usepackage{graphicx}
\usepackage{textcomp}
\usepackage{xcolor}
\usepackage{makecell}
\usepackage{fancyhdr}
\begin{document}
%tweaks to enable paper to fit into 8 pages:
\bstctlcite{IEEEexample:BSTcontrol}

\title{Hierarchical Multi-Process Fusion for Visual Place Recognition}

\author{Stephen Hausler and Michael Milford
\thanks{
The work of S. Hausler was supported by a Research Training Program Stipend. The work of M. Milford was supported by the Australian Research Council Future Fellowship under Grant FT140101229.
The authors are with the ARC Centre of Excellence for Robotic Vision, Queensland University of Technology, Brisbane, QLD 4000, Australia (e-mail: stephen.hausler@hdr.qut.edu.au).}
}

\maketitle
\thispagestyle{fancy}

\pagestyle{empty}

\begin{abstract}
Combining multiple complementary techniques together has long been regarded as a way to improve performance. In visual localization, multi-sensor fusion, multi-process fusion of a single sensing modality, and even combinations of different localization techniques have been shown to result in improved performance. However, merely fusing together different localization techniques does not account for the varying performance characteristics of different localization techniques. In this paper we present a novel, hierarchical localization system that explicitly benefits from three varying characteristics of localization techniques: the distribution of their localization hypotheses, their appearance- and viewpoint-invariant properties, and the resulting differences in where in an environment each system works well and fails. We show how two techniques deployed hierarchically work better than in parallel fusion, how combining two different techniques works better than two levels of a single technique, even when the single technique has superior individual performance, and develop two and three-tier hierarchical structures that progressively improve localization performance. Finally, we develop a stacked hierarchical framework where localization hypotheses from techniques with complementary characteristics are concatenated at each layer, significantly improving retention of the correct hypothesis through to the final localization stage. Using two challenging datasets, we show the proposed system outperforming state-of-the-art techniques.

\end{abstract}

\section{Introduction}

A hierarchical approach to localization is a well-established process with roots in computational efficiency and provides a method of improving spatial accuracy of localization. Hierarchies have also been discovered in the mammalian brain, both in the structure of grid cells in the Hippocampus \cite{Stensola2012}, and in the visual pathway of the Visual Cortex \cite{Kruger2013}. In this research we ask the question: does a hierarchical approach to visual localization provide a direct improvement to the localization success rate when different image processing methods are used, and how should such a hierarchical approach be structured? To answer this question, we perform an extensive investigation into combining different combinations of local, global and deep learnt image descriptors in a  hierarchical localization pipeline. We showcase our findings using the datasets Nordland and Berlin Kurfurstendamm. The Nordland dataset tests our hierarchical fusion under severe appearance change but no viewpoint change, while Berlin verifies these results under severe viewpoint change.

\begin{figure}[h]
\centering
\includegraphics[width=\linewidth,trim=0.1cm 0.5cm 0.5cm 4cm,clip]{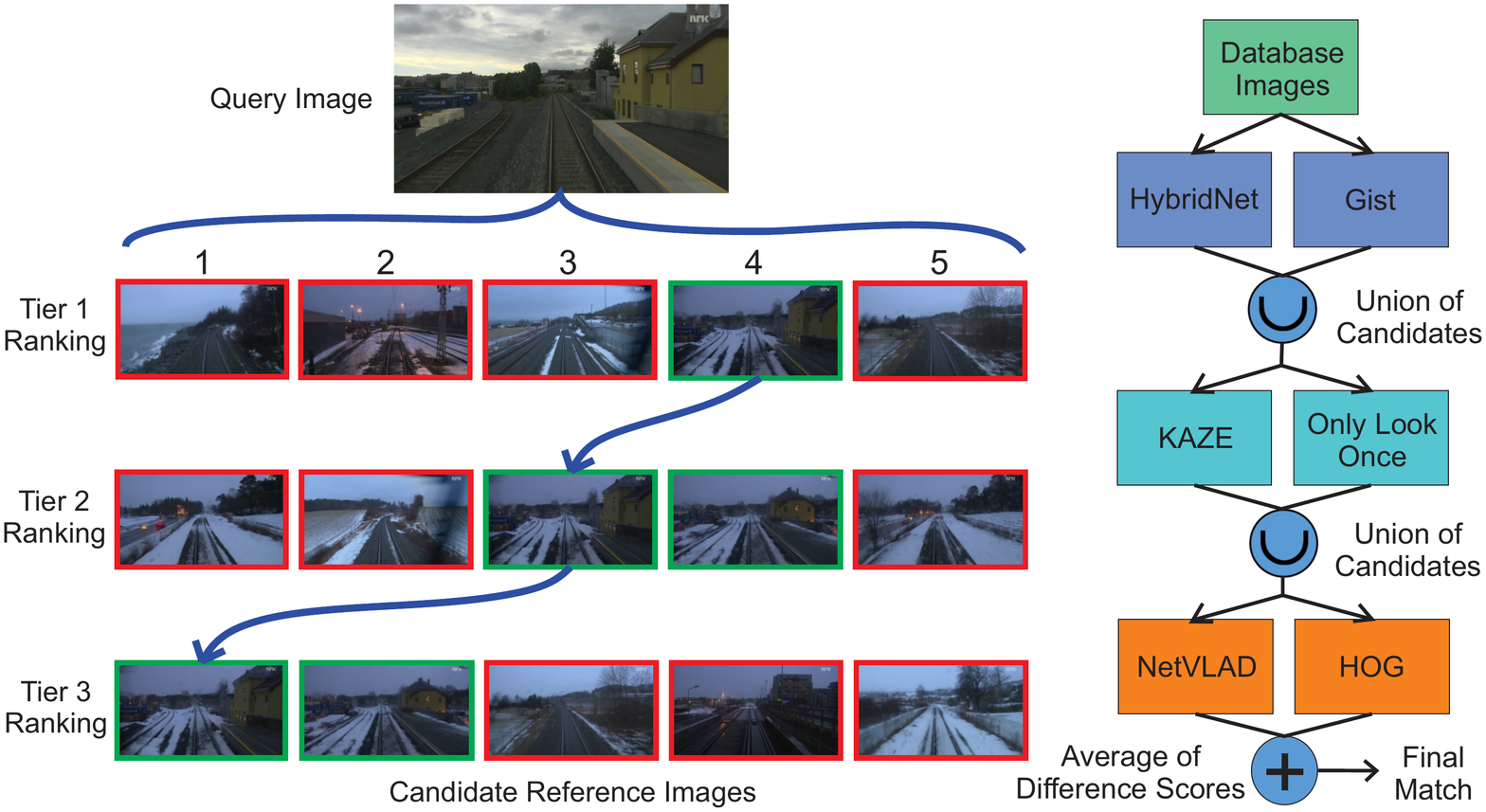}	%trim left,bottom,right,top  
\caption{Our approach selects best-matching candidate reference images for different tiers in a hierarchical approach. The top candidates in Tier 1 are passed to Tier 2, which finds a smaller number of top candidates to pass to Tier 3. The top ranked image in Tier 3 is the location we loop close to. As we move down the hierarchy, the ground-truth match is successfully shifted to the front of the ranked list, by virtue of the use of different but complementary image processing methods in each tier. The candidate ranking example shown is an experimental result from the Nordland dataset and a green border denotes a matching image that is within the ground-truth tolerance for our experiments.}
\label{Match_Example}
\end{figure}

In this paper we provide the following contributions:
\begin{itemize}
\item{We show how merely fusing together multiple visual place recognition techniques in parallel is inferior to a hierarchical approach.}
\item{We develop a hierarchical framework that can be configured to improve both computation speed and localization performance, and demonstrate its superior performance using both two and three tier architectures.}
\item{We show how the individual performance of a place recognition method does not always directly predict its utility in a hierarchical system, and show how combining different techniques can result in superior performance compared to stacking a single, higher performing technique.}
\item{We expand the system to enable concatenation of multiple place recognition techniques within a level of the hierarchy, leading to improved retention of the correct place recognition hypothesis that results in additional improvements in performance.}
\end{itemize}

The source code for this paper is available online\footnote{https://github.com/StephenHausler/Hierarchical-Multi-Process-Fusion}. 

\section{Related Work}

In solutions to solve the data association problem in localization (perform loop closure), a hierarchical approach is commonly used \cite{Chen2015,Lost2018,Sarlin,Garcia-Fidalgo2017,Korrapati2014,Maohai2011,Mohan2015}. In many of these approaches, a coarse global search is performed across all stored database image representations before a second, fine-grained search is used to filter the set of candidates produced by the global search \cite{Lost2018}. Often the second search will utilise computationally-intensive geometric approaches, such as Bundle Adjustment \cite{Triggs2000} and Co-visibility Clustering \cite{Sarlin}.

In these hierarchies, a wide-variety of image descriptors have been used. These descriptors fall into the categories of global \cite{DN2005, Oliva2001}, local \cite{Alcantarilla2012,BH2008} and deep learnt \cite{Arandjelovic2018,ConvNet}. Commonly a different category of descriptor is used in different stages of a hierarchical approach. Maohai et. al. performed a two-stage localization hierarchy, using a color histogram global image descriptor to provide a coarse localization and then evaluated the resultant candidates using SIFT feature matching \cite{Maohai2011}. A more advanced version of this was developed which uses the PHOG descriptor at the first stage of the hierarchy, then uses FAST corners with LDB binary descriptors \cite{Garcia-Fidalgo2017}. Their work also included RANSAC verification and a Bayes Filter to further improve localization. Prior work has also investigated the combination of deep-learnt and local features in a hierarchy, showing that accurate 6-DoF pose estimates can be produced at real-time if the set of candidates is first filtered using a deep-learnt global descriptor \cite{Sarlin}. 

A single image descriptor can also be used in a hierarchy. One approach is to sub-analyze the images, for example, using patch-verification and Sum-of-Absolute Differences \cite{Milford2014}. Alternatively, a sequence of images can be interpreted across multiple spatial scales, providing a hierarchical set of multi-scale clustered descriptors of the current scene \cite{Chen2015}.

While many of the aforementioned localization approaches have used one or two image processing methods in their hierarchy, none of those approaches use an arbitrarily large number of methods. Fusing a large number of image descriptors has had some related investigation, such as SRAL \cite{SRAL}, which simultaneously used six different types of visual features. In the author's previous work, four different image processing methods were fused in a temporal sequence \cite{Hausler2019}. Neither of these two approaches utilised a hierarchical framework, while this work showcases using multiple image processing methods in a hierarchy.

\section{Proposed Approach}

In this work we present a sequence of investigations into hierarchical fusion of place recognition techniques which inform subsequent design of a novel, high performing hierarchical place recognition framework. In a typical hierarchical localization system, loop closure candidates from a first, computationally cheap localization method are used to define a set of potential matches for evaluation by a second, computationally expensive sub-system. This pipeline allows for efficient real-time localization, even in long-term navigation trials. In our experiments we use a three-tier hierarchy, however, our proposal is customizable and can be applied to any arbitrary number of tiers. Each tier uses a different image processing method to evaluate the similarity between the currently viewed scene and the provided candidates. Additionally, we include the option of adding additional image processing methods within a particular tier \cite{Garg2019}, such that the selected candidates from that tier becomes the union of the best candidates from the multiple methods within that tier. In this section, we will begin by describing the configuration of each image processing method we use in our experiments.

\subsection{Design and Configuration of Image Processing Methods}

For this work, we selected a total of six different image processing methods (half hand-crafted, half deep learnt). Our proposed approach is equally applicable to methods not chosen and any of the following methods can be replaced with an alternate approach.

\textbf{Histogram of Oriented Gradients (HOG) -} we use Dalal and Trigg's HOG \cite{DN2005} with a cell size of 30 by 30 pixels. We also re-size the input images to 300 by 300 pixels, which assists in alleviating small appearance variations while also reducing the size of the feature vector produced by HOG.

\textbf{Gist -} uses Gabor filters to extract gradients from an image, for a range of spatial scales and frequencies \cite{Oliva2001}. Using the default settings, Gist outputs a 512 dimension feature vector from an input image. 

\textbf{KAZE -} is a local feature detector and descriptor similar to SURF \cite{BH2008} or SIFT \cite{Lowe2004}, except it has demonstrated improved feature quality but is also computationally expensive \cite{Alcantarilla2012}. We match features between two images using MATLAB's built-in \emph{matchFeatures} function, and specify a match filter with a \emph{MatchThreshold} of 20 and a \emph{MaxRatio} of 0.7. By applying the filter, we remove incoherent matches which fail Lowe's ratio test \cite{Lowe2004}. The distance between a query and a database image is then the sum of the residual distances between the twenty strongest matching features. The database images with the smallest distance are considered the best matching candidates. 

\textbf{NetVLAD -} is a neural network designed for place recognition, inspired by the success of VLAD \cite{Arandjelovic2018}. We use the network pre-trained on Pittsburgh 30k and re-size our images to fit the input size of the network. We match images using the computed NetVLAD feature vector, which has a dimensionality of 4096. 

\textbf{HybridNet -} is a re-trained version of AlexNet, trained on images recorded by a collection of security cameras over an extended period of time \cite{CZ2017}. In our use of this network, we extract a feature vector from the Conv5 layer and use an aggressive method of dimensionality reduction. We compose the feature vector by aggregating the spatial (W x H) position of non-zero maximum activations across all the feature maps. As W and H are both dimension 13 in Conv5, this method produces a feature of dimension 169.

\textbf{Only Look Once -} in this method, later convolutional layers are used to find spatial regions with the strongest activations \cite{LookOnce2017}. Multiple region descriptors are then created from the activations within each of these spatial regions in an earlier convolutional layer. To calculate the similarity between two images, these region descriptors are cross matched. This image processing method can match images across both viewpoint and appearance variations, but it is computationally expensive. We use the open-source version of Only Look Once\footnote{https://github.com/scutzetao/IROS2017\_OnlyLookOnce} to calculate the image similarity score.

\subsection{Computation of Normalized Difference Scores}

Each aforementioned image processing method produces a difference score between the current query image and either every database image or the set of candidates in the previous tier of the hierarchy. These raw difference scores have a wide variation in their data spread. Therefore we use min-max normalization to normalize all difference scores to the range of 0 to 1, where 1 denotes the best matching database image and 0 the worst.
\begin{equation}
    Dnorm = \frac{D - \text{max}(D)}{\text{min}(D) - \text{max}(D)}
\end{equation}

\subsection{Fusion of Multiple Methods in a Hierarchy}

The first tier of our hierarchy performs a global search across all database images, and returns $k_{t1}$ nearest neighbour candidates with respect to the current query image. If multiple image processing methods are used in the first tier, then the returned candidates are the union of the nearest neighbours from each method in tier 1:
\begin{equation}
    C(k_{t1}) = C(k_{m1}) \cup C(k_{m2}) .. \cup C(k_{mn})
\end{equation}
\noindent where $m_1 .. mn$ are the methods from tier 1, up to $n$ methods in this tier. $C$ denotes the set of candidates, with a number of candidates up to $k$.

Candidates $C(k_{t1})$ are then passed to the second tier of the hierarchy, to be evaluated by a more fine-grained search across this smaller set of `potentially good' candidates. The image processing methods in tier 2 can and likely should be different to those in tier 1, with characteristics that enable the differentiation of perceptually aliased candidates. Because tier 2 only has to analyze a small number of candidates, rather than the entire database, the methods used can be more computationally intensive. $k_{t2}$ nearest neighbour candidates are selected from this tier, comparing each candidate to the current query image, with the formulation described in Equations 3 and 4.
\begin{equation}
    k_{t2} < k_{t1}
\end{equation}
\begin{equation}
    C(k_{t2}) = C(k_{m1}) \cup C(k_{m2}) .. \cup C(k_{mn})
\end{equation}

Methods $m$ in tier 2 are different to the $n$ methods in tier 1 and the values of $n$ can be different or the same between tiers.

At this point, further tiers can be added as needed, continuing to pass a shrinking pool of candidates. However, once the final tier of the hierarchy is reached, a best match consensus is determined. As the best match is a singular value, the union operator can no longer be applied between different image processing methods in the one layer. Instead, we calculate the mean normalized difference score across these multiple methods. The largest mean scoring candidate is then selected as the best match from the final tier of the hierarchy, as described by Equations 5 and 6.
\begin{equation}
    D_{t3} = \frac{D(k_{t2})_{m1} + D(k_{t2})_{m2} + .. D(k_{t2})_{mn}}{n}
\end{equation}
\begin{equation}
    bestCand = \text{argmax}(D_{t3})
\end{equation}

With a selection of complementary image processing methods, accurate localization can generally be achieved at this final tier. However, an edge case can exist where the earlier layers successfully identify the global best match candidate while the later layer, with a different image processing method, is unable to identify the correct match. To guard against this condition, we provide an extension which fuses the difference scores from the best matching candidates in earlier layers to the final layer decision process.

\subsection{Enhancing Localization using Earlier Layers}

To further improve localization, the mean normalized difference from the final tier is added to difference scores from those same candidates in earlier tiers. Assuming a three tier hierarchy, the final tier will produce $k_{t2}$ normalised difference scores for a list of $C(k_{t2})$ candidates. The difference scores from earlier layers are then extracted for the final tier candidate set, resulting in $k_{t2}$ scores per layer. To make scoring equivalent and hence combinable across tiers, we re-normalize the extracted difference scores to fall in the range of 0 to 1. Because earlier tier methods may be worse-performing comparing to later tier methods, we include the option of biasing the summation (making it less fair) using pre-calibrated weight scalars for each tier:
\begin{equation}
    D_{final} = D_{t3}W_3 + D_{t2}W_2 + D_{t1} W_1
\end{equation}
\noindent where $D_{t2}$ and $D_{t1}$ are the normalized subset of all difference scores, as described by Equations 8-10. For all our experiments, we set the weight scalers to the values 1, 0.75 and 0.5. Our choice is based on the observation that the Recall at 1 performance is generally better in the later tiers than in the earlier tiers (see Section IV.).

Because the candidates passed to later layers are the concatenation of candidates from each method in the same tier, we want to use the maximum difference scores to guarantee that only the best performing method in a layer is being used in the final calculation. We begin by finding the maximum difference score for each candidate id across the different methods:
\begin{equation}
    D_{t2} = \text{max}(D_{m1},D_{m2},..,D_{mn})
\end{equation}
\begin{equation}
    D_{t2} = D_{t2}(i \in C(k_{t2})) \; \text{for} \; i = 1:\text{size}(D_{t2})
\end{equation}
\begin{equation}
    D_{t2} = \frac{D_{t2} - \text{min}(D_{t2})}{\text{max}(D_{t2}) - \text{min}(D_{t2})}
\end{equation}

Equations 8-10 are repeated for $D_{t1}$ and any other tiers prior to the last tier.

The final step involves performing normalization by standardizing the data to have mean 0 and standard deviation of 1.
\begin{equation}
    D_{final} = \frac{D_{final} - \mu(D_{final})}{\sigma(D_{final})}
\end{equation}

The best matching candidate is then the maximum score in $D_{final}$. In our results, we call this the \emph{Combined} recall.

\section{Results}

\subsection{Dataset Configuration}

We evaluate our proposal using the publicly available and widely used datasets Nordland \cite{NordlandDatasetRef} and Berlin Kurfurstendamm \cite{Suenderhauf2015}. These two datasets capture a range of relevant place recognition challenges with significant appearance variation on Nordland and large viewpoint shifts on Berlin. We split each dataset into train and test sets and we use the training set to evaluate different combinations of image processing methods in our hierarchical approach.

The Nordland dataset consists of a 728 km train trip through Norway, across four different seasons. In our experiments we use the Summer and Winter seasons, where our database contains Winter images while our query set is from Summer. To generate our training and test sets, we extracted frames at 1 FPS from the original videos, omitting sections where the train is either stopped or in a tunnel \cite{ConvNet}. Our training set contains 1000 frames extracted from the start of the Nordland train trip. Our test set also contains 1000 frames, except these frames are taken from a later section of the train route. For all experimental results on Nordland we use a ground-truth tolerance of 10 frames. 

The Berlin Kurfurstendamm dataset contains a collection of images downloaded from Mapillary \cite{Sunderhauf2015}, captured in the city of Berlin along the road Kurfurstendamm. For our training set, we use 280 images recorded from a bicycle as our query set. Our reference set contains 314 images captured by both a car and a bus driving on the same road. Our test set is similar, except the query images are recorded by a different bicycle, travelling on the same road on a different date with several years time gap between the two query sets. We use a ground-truth tolerance of 50 meters, since there is a large real-world distance between successive frames. 

\subsection{Evaluate Individual Methods on Train Set}

\begin{figure}[h]
\centering
\includegraphics[width=0.9\linewidth,trim=0cm 3.7cm 0cm 0cm,clip]{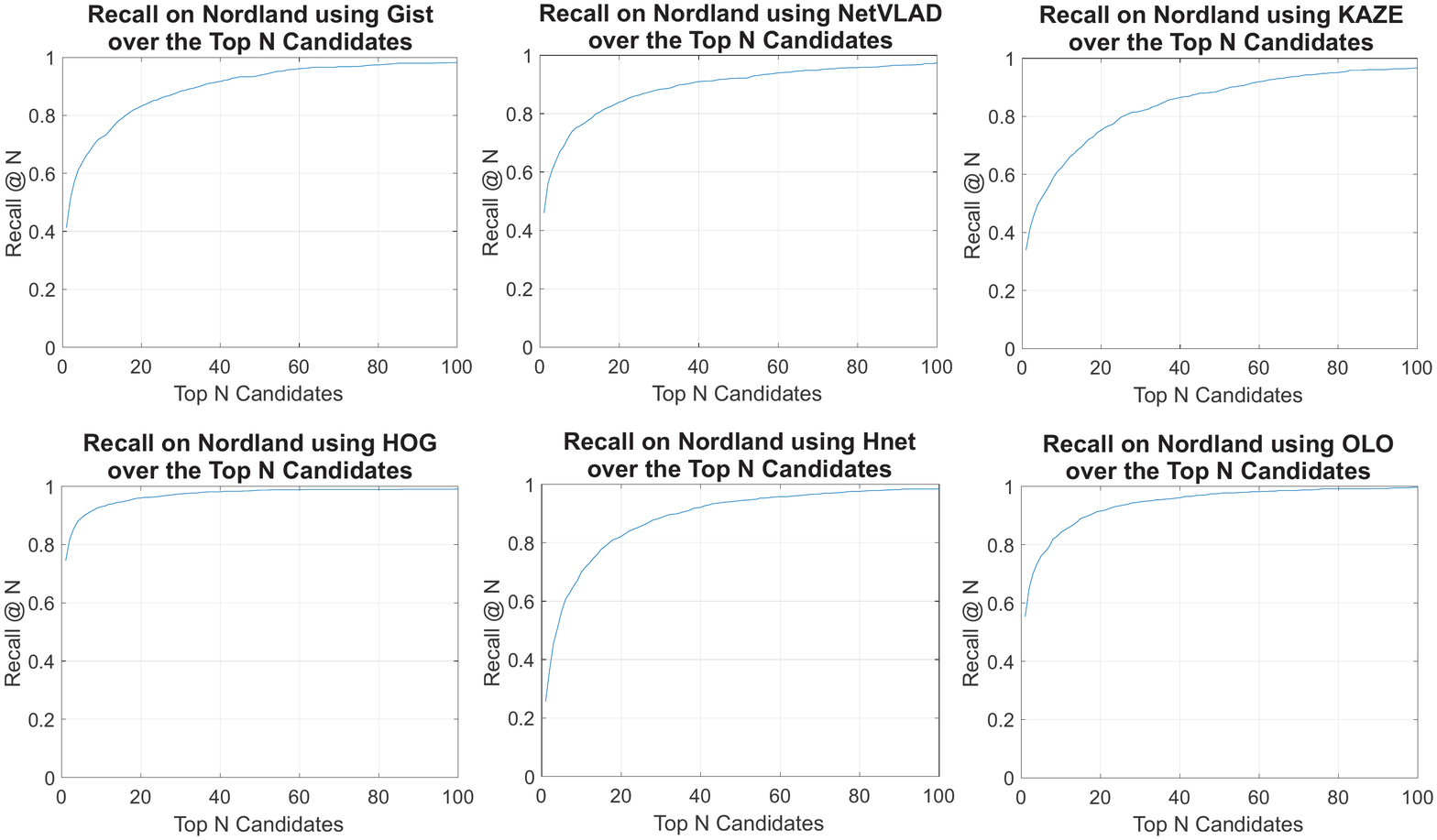}	%trim left,bottom,right,top  
\caption{Recall @ N curves for Gist, NetVLAD, KAZE, HOG, HybridNet and Only Look Once (OLO) on the Nordland Trainset. The recall @ 1 performance varies significantly between methods, however the recall @ 10 tends to be more consistent between methods.}
\label{Charact1}
\end{figure}

\begin{figure}[h]
\centering
\includegraphics[width=0.9\linewidth,trim=0cm 3.7cm 0cm 0cm,clip]{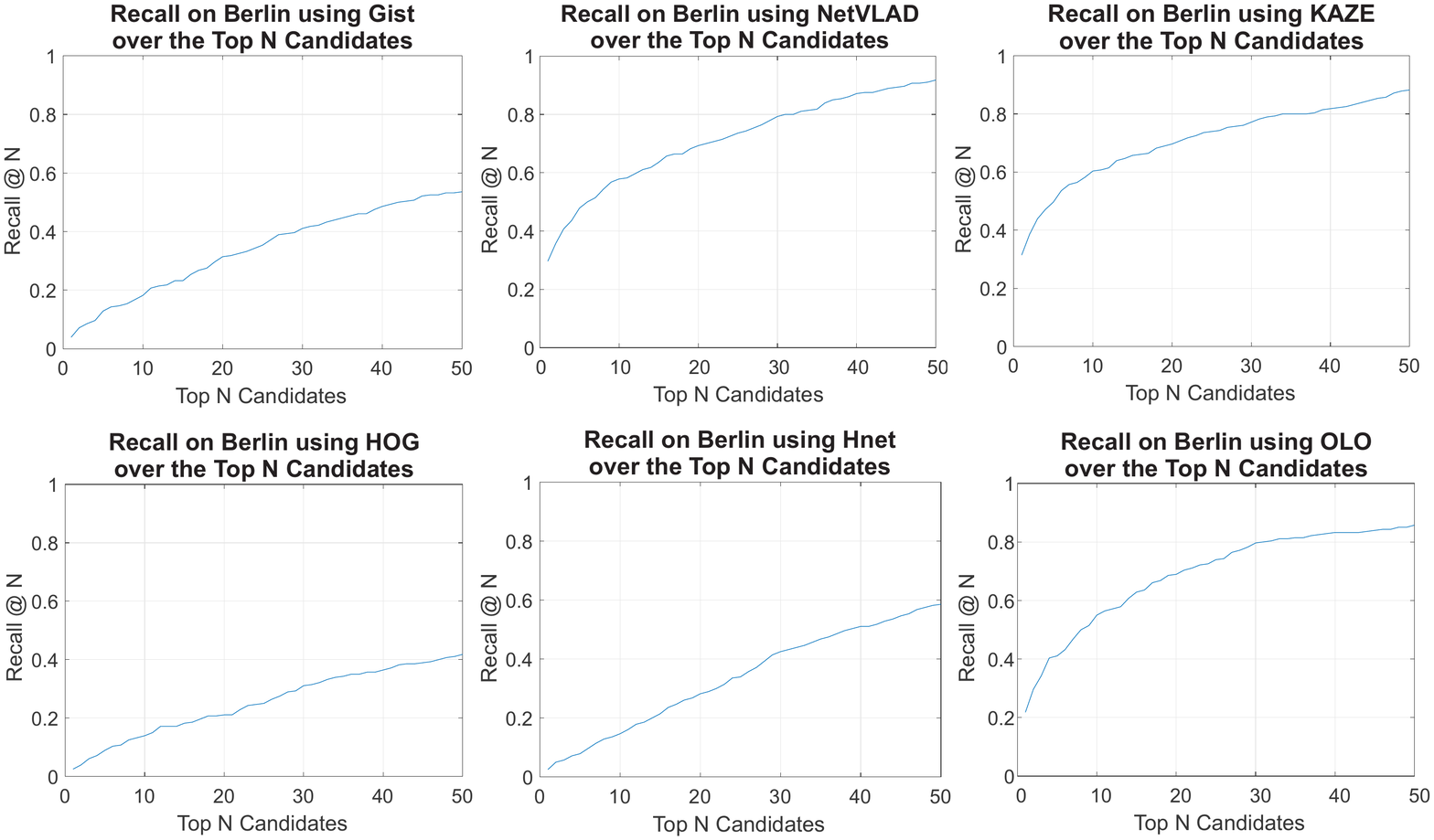}	%trim left,bottom,right,top
\caption{Recall @ N curves for Gist, NetVLAD, KAZE, HOG, HybridNet and Only Look Once (OLO) on the Berlin Trainset. NetVLAD, KAZE and OLO perform consistently well compared to HOG, Gist and HybridNet, which cannot handle the large viewpoint changes.}
\label{Charact1_Berlin}
\end{figure}

To begin our experiments, we analyze the performance of each individual image processing method on the two training datasets (see Figures \ref{Charact1} and \ref{Charact1_Berlin}). We display the localization using the Recall @ N metric, where N is the number of top candidates. As more top candidates are used, the recall approaches one as the ground-truth matching candidate is more likely to exist in the larger set of candidates. We limit N to 50 for Berlin and 100 for Nordland, since the Berlin dataset is significantly smaller than Nordland. Different visual features exhibit substantially different Recall @ N characterizations. Additionally, some methods are more suited to a particular dataset. For example, HOG has outstanding performance on Nordland, where there is no viewpoint changes, but localizes poorly on Berlin (which has very large viewpoint shifts).

\subsection{Combine Two Methods in a Hierarchy}

\begin{figure}[p]
\centering
\includegraphics[width=0.8\linewidth,trim=0cm 3.7cm 0cm 0cm,clip]{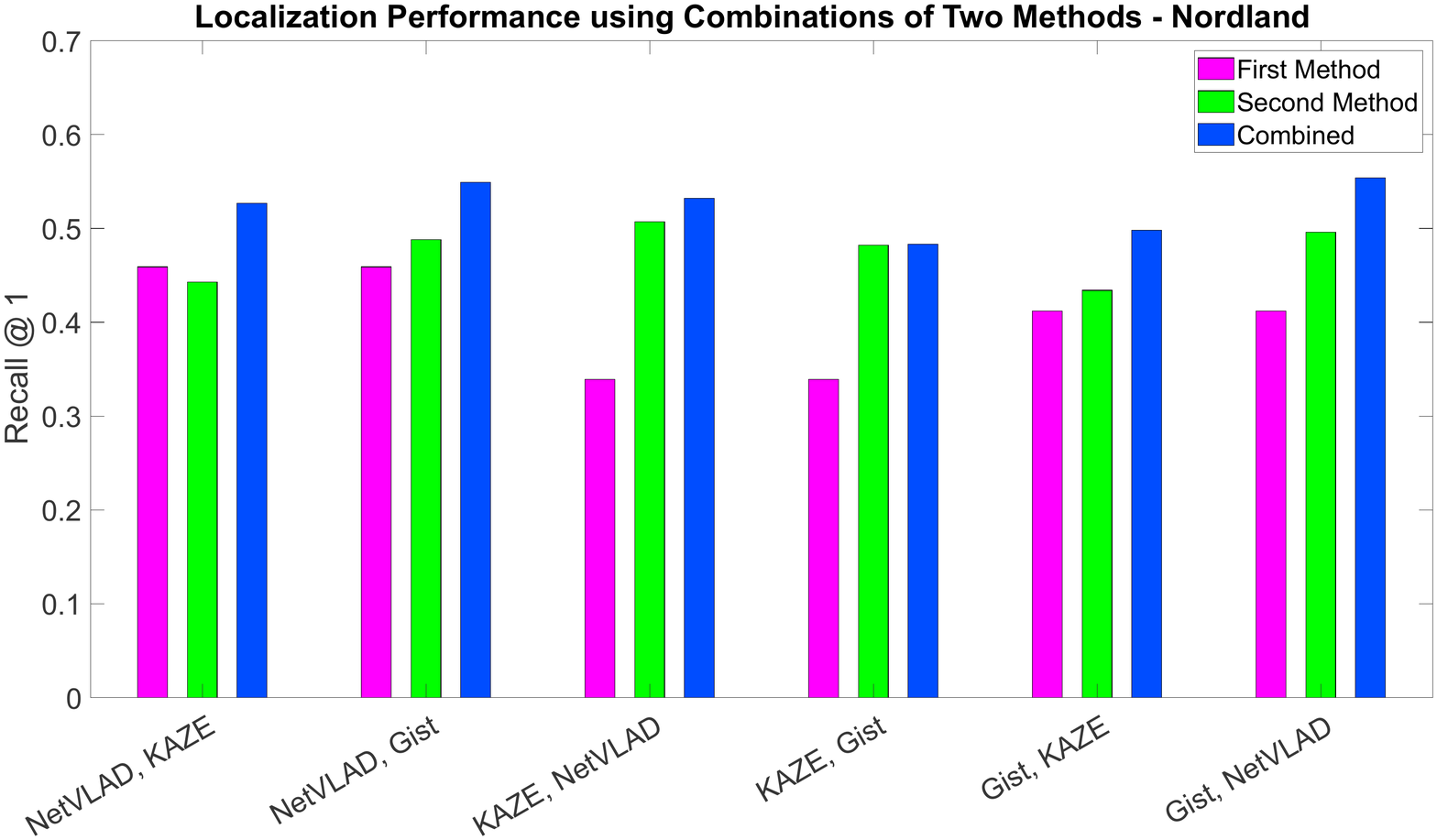}	%trim left,bottom,right,top  
\caption{For the Nordland training set, we combine different sets of two methods and show the recall at the top candidate. For all three methods, if the same method is used in the second tier of the hierarchy, the recall @ 1 improves. Any order of methods in the hierarchy improves the Combined localization rate.}
\label{TwoMeth}
\end{figure}

In this experiment, we combine two different localization techniques in a hierarchy, where the first method passes the top 50 (Berlin) or 100 (Nordland) candidates to the second method. The second method only has to select the best candidate out of 50 or 100 potentially good candidates, rather than selecting the best candidate out of the full reference database. For the Nordland train set, we select two methods out of Gist, NetVLAD and KAZE and exhaustively evaluate all combinations of these methods (Figure \ref{TwoMeth}). We chose Gist, NetVLAD and KAZE in order to combine a global descriptor, a deep-learnt approach, and a local feature detector. Additionally, these three methods exhibited similar Recall @ N characterizations, thus providing the fairest analysis of the real benefits of combining multiple methods in a hierarchy. 

\begin{figure}[p]
\centering
\includegraphics[width=0.8\linewidth,trim=0cm 3.7cm 0cm 0cm,clip]{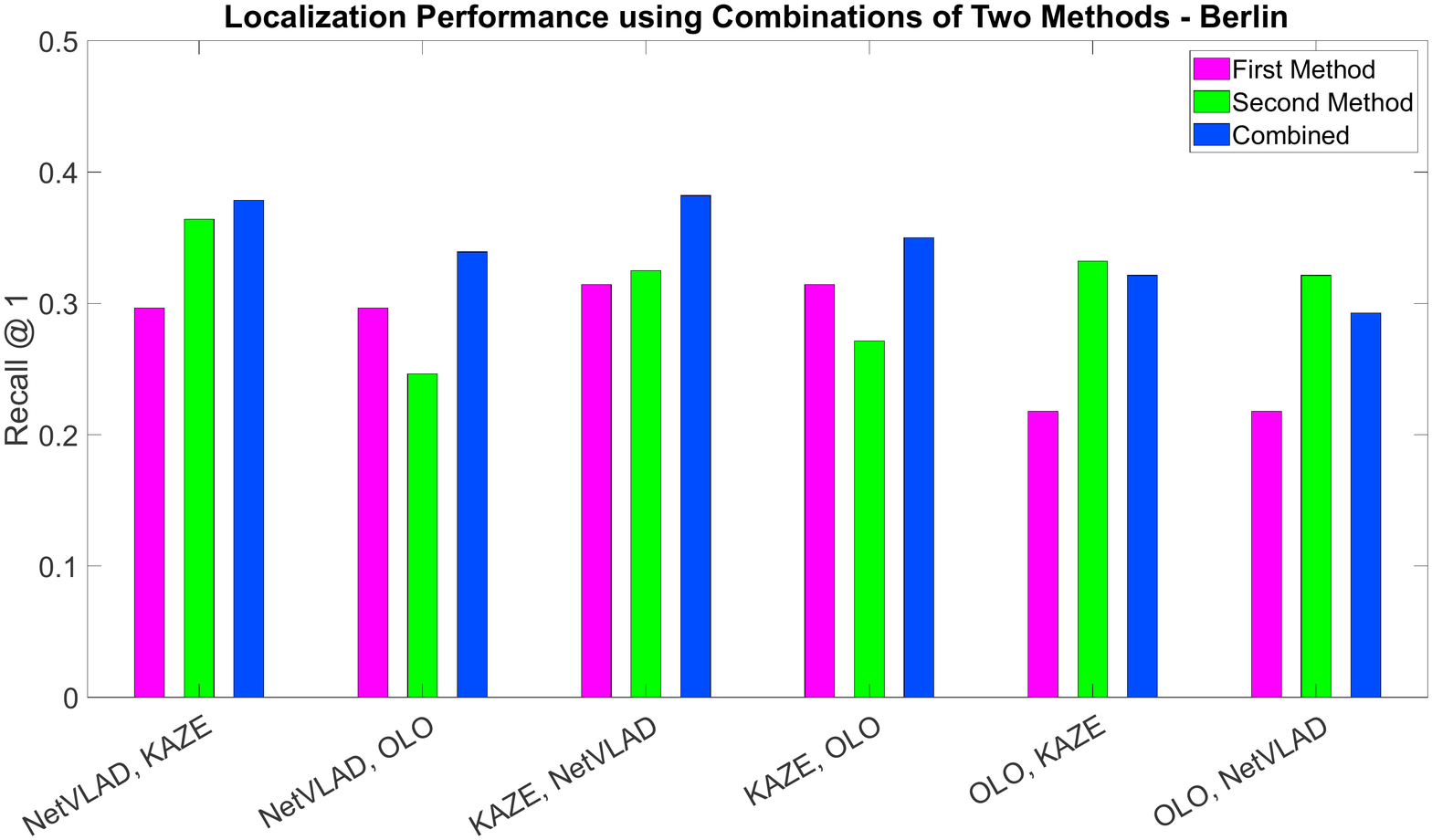}	%trim left,bottom,right,top  
\caption{For the Berlin dataset, again we combine different sets of two methods and show the recall at the top candidate. Only Look Once is the worst performing method and cannot compete with the combination of NetVLAD and KAZE.}
\label{TwoMeth_Berlin}
\end{figure}

We use the methods NetVLAD, KAZE and OLO when evaluating the Berlin training set (Figure \ref{TwoMeth_Berlin}), because of the poor performance of the other three methods (as determined in Section IV. B.). We show our results using the recall at the best candidate for each method individually and for the final combination (using the algorithm described in Section III. D.).

\begin{figure}[p]
\centering
\includegraphics[width=0.85\linewidth,trim=0cm 4.3cm 0cm 0cm,clip]{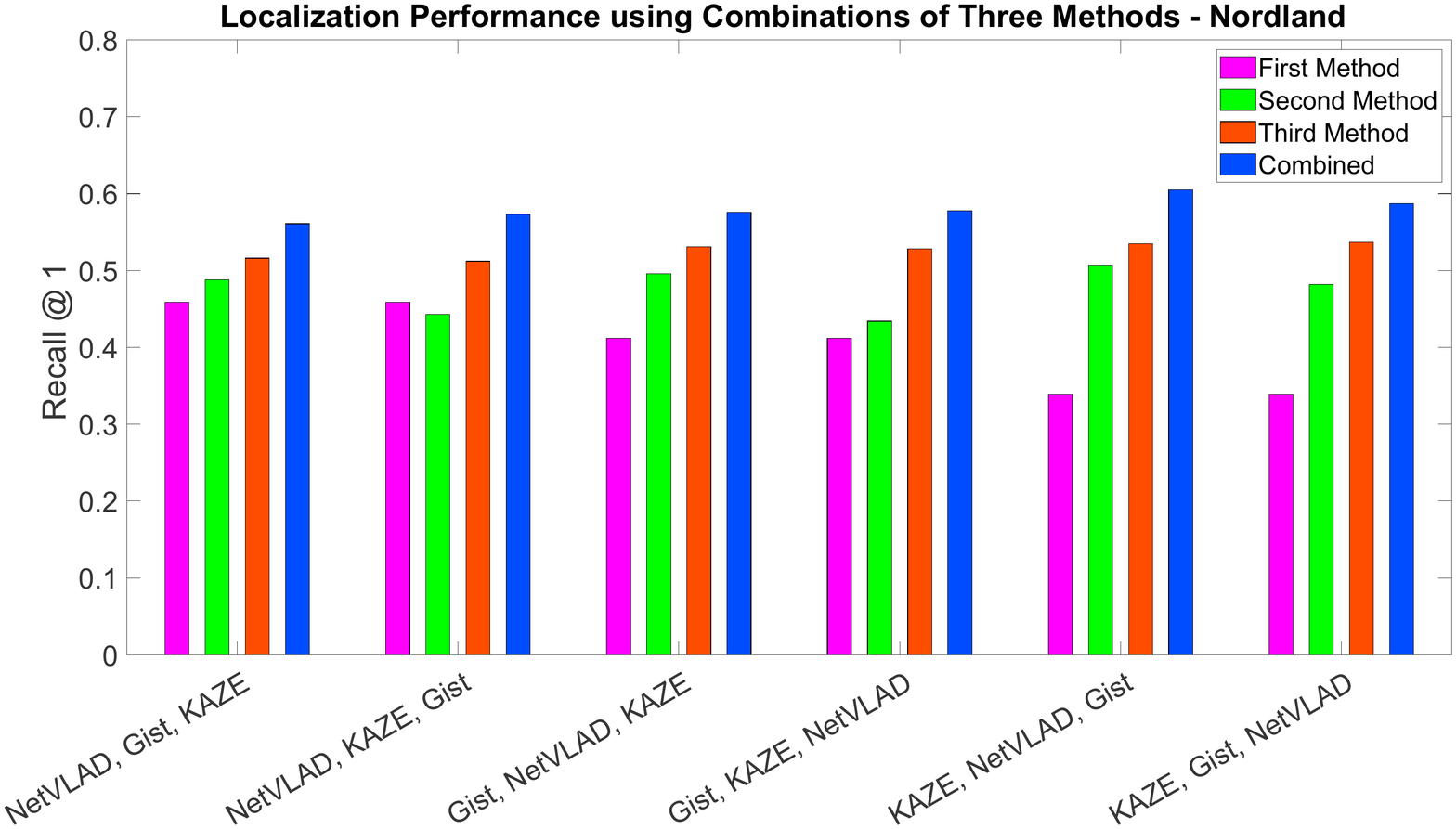}	%trim left,bottom,right,top  
\caption{We combine three different methods and show the recall at the top candidate on the Nordland train set. An increasing trend exists, where the recall @ 1 improves as the hierarchy is progressed and the Combined recall produces the best result.}
\label{ThreeMeth}
\end{figure}

\begin{figure}[p]
\centering
\includegraphics[width=0.85\linewidth,trim=0cm 4.3cm 0cm 0cm,clip]{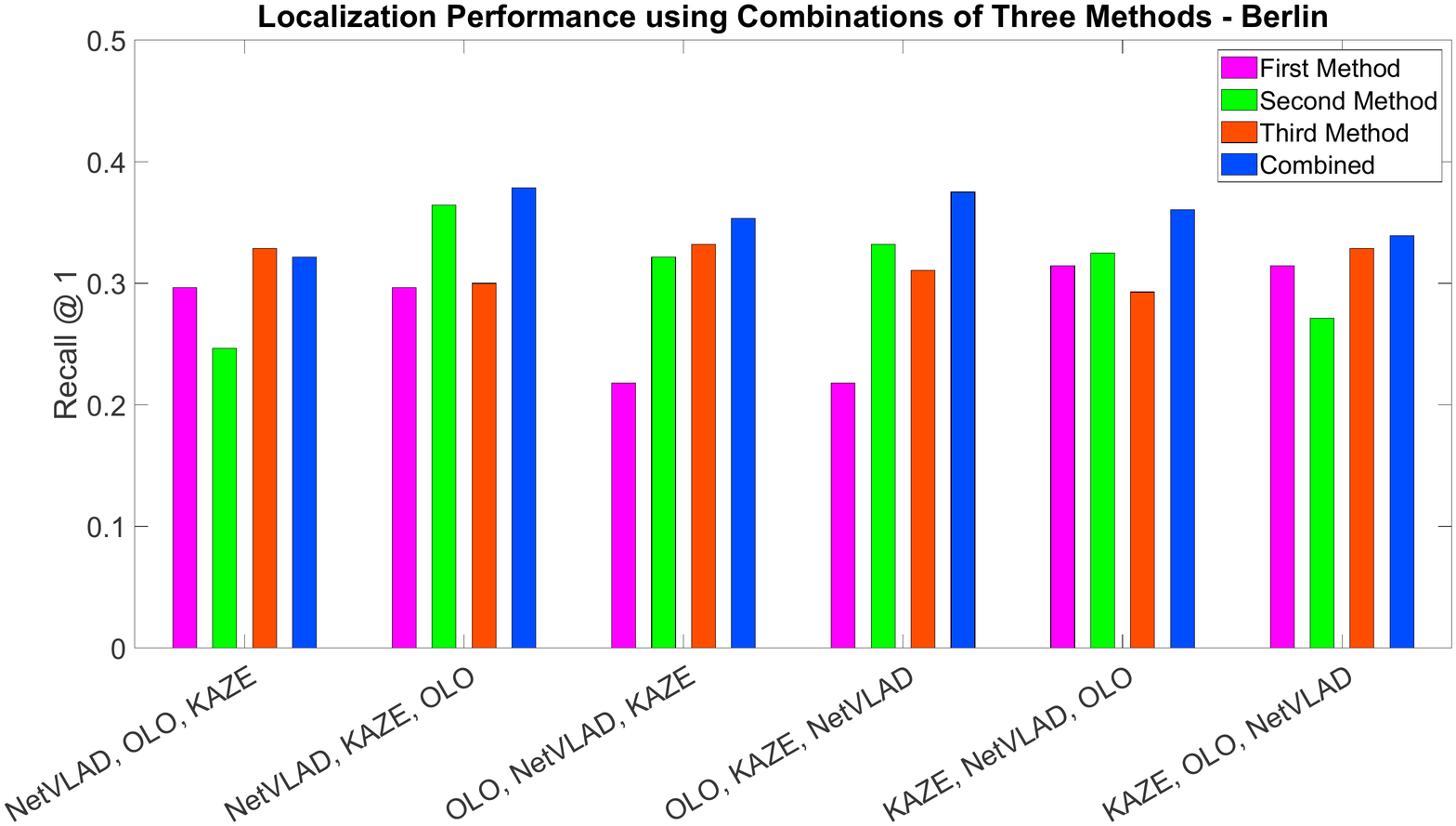}	%trim left,bottom,right,top  
\caption{For the Berlin training dataset, we combine three different methods and show the recall at the top candidate. Because Only Look Once is less suited to this training dataset, using the two methods NetVLAD and KAZE has the same Recall @ 1 as using three methods.}
\label{ThreeMeth - Berlin}
\end{figure}

\subsection{Combine Three Methods in a Hierarchy}

By adding a third tier to the hierarchy, we can pass a small number of potential best candidates to a method which can distinguish the best match from a small number of perceptually aliased images. We again pass 50 or 100 candidates from the first tier to the second method, then we pass 10 candidates from the second tier to the final method. In Figure \ref{ThreeMeth} for the Nordland dataset, the Recall @ 1 consistently improves as the number of candidates is reduced, irrespective of the order of the methods. The mean Recall @ 1 for the First, Second and Third Methods is 40.3\%, 47.5\% and 52.7\% respectively, and 58.0\% for the Combined.

Figure \ref{ThreeMeth - Berlin} (Berlin trainset) reveals that the benefits of the hierarchical approach are not just because of inherent benefits of a shrinking candidate pool, but also the complementary interactions between different methods. For example, the Recall @ 1 for KAZE in the Second Method position is 36.4\% after receiving candidates from NetVLAD, while the Recall @ 1 for KAZE after receiving candidates from Only Look Once is 33.2\%. The mean Recall @ 1 for the First, Second and Third Methods is 27.6\%, 31.0\% and 31.6\% respectively, and 35.5\% for the Combined.  

\subsection{Combine Multiple Methods in a Single Tier}

In this experiment, we maintain the same number of tiers except now we have two methods within each tier. Using two methods per tier increases the retention of the correct place recognition hypothesis if environment variations cause a particular method to perform poorly. We use the original four methods showcased in the previous experiments, except we also the additional methods HybridNet and HOG. Out of these six methods, we paired methods together based on both the length of the feature vector produced and the type of algorithm. For example, HybridNet (Hnet) and Gist have the smallest feature vectors (169 and 512 respectively), while NetVLAD and HOG use vectors of size 4096 and 2916. Only Look Once and KAZE both do not produce a feature vector and instead have unique image comparison algorithms.

\begin{figure}[h]
\centering
\includegraphics[width=0.8\linewidth,trim=0cm 4.4cm 0cm 0cm,clip]{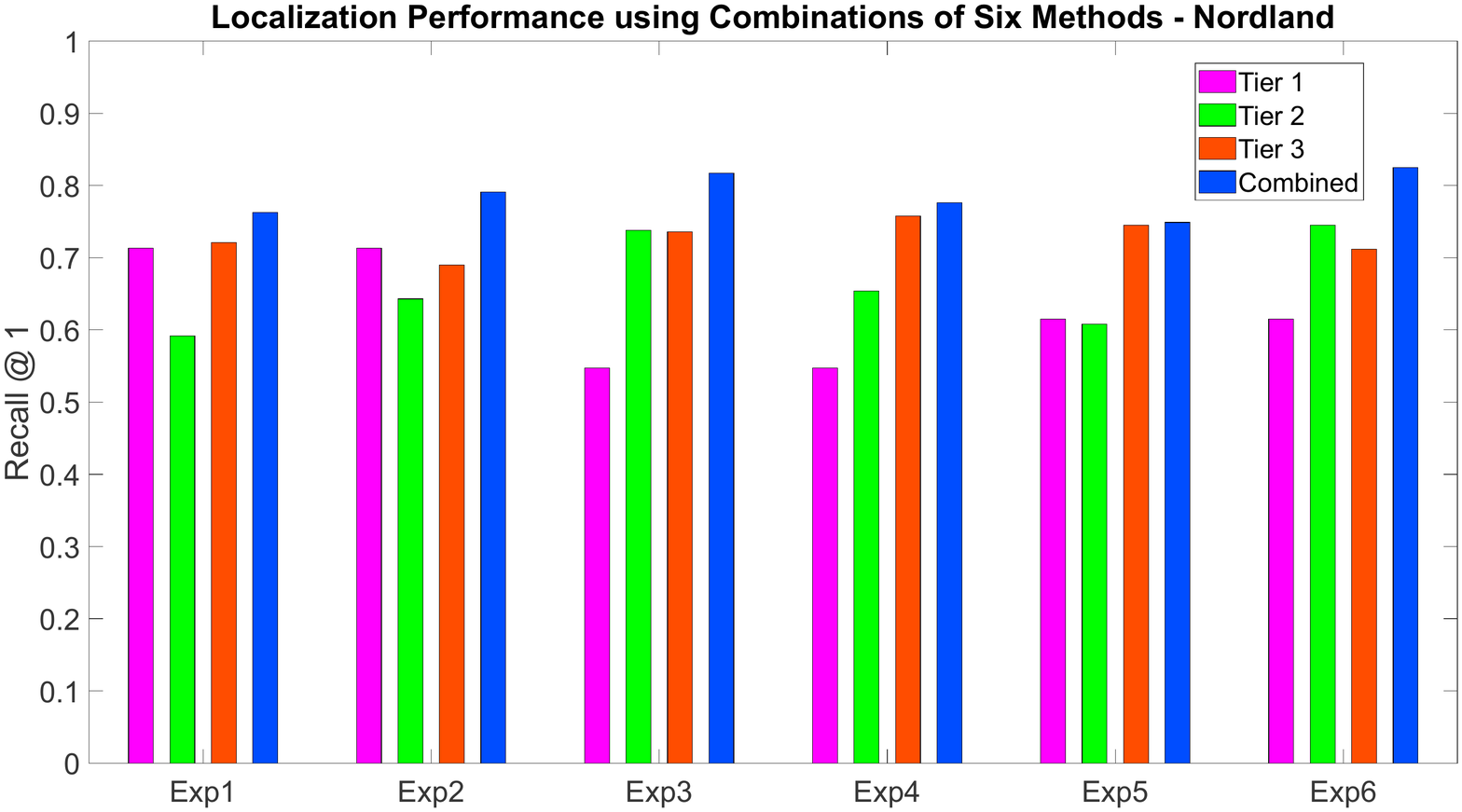}	%trim left,bottom,right,top  
\caption{Results for 6 methods on Nordland, where there are two methods per tier of the hierarchy. When 6 methods are used, because the first two tiers concatenate the candidates from each method, 200 and 20 candidates are passed from their respective tiers.}
\label{SixMeth}
\end{figure}

\begin{figure}[h]
\centering
\includegraphics[width=0.82\linewidth,trim=0cm 4.7cm 0cm 0cm,clip]{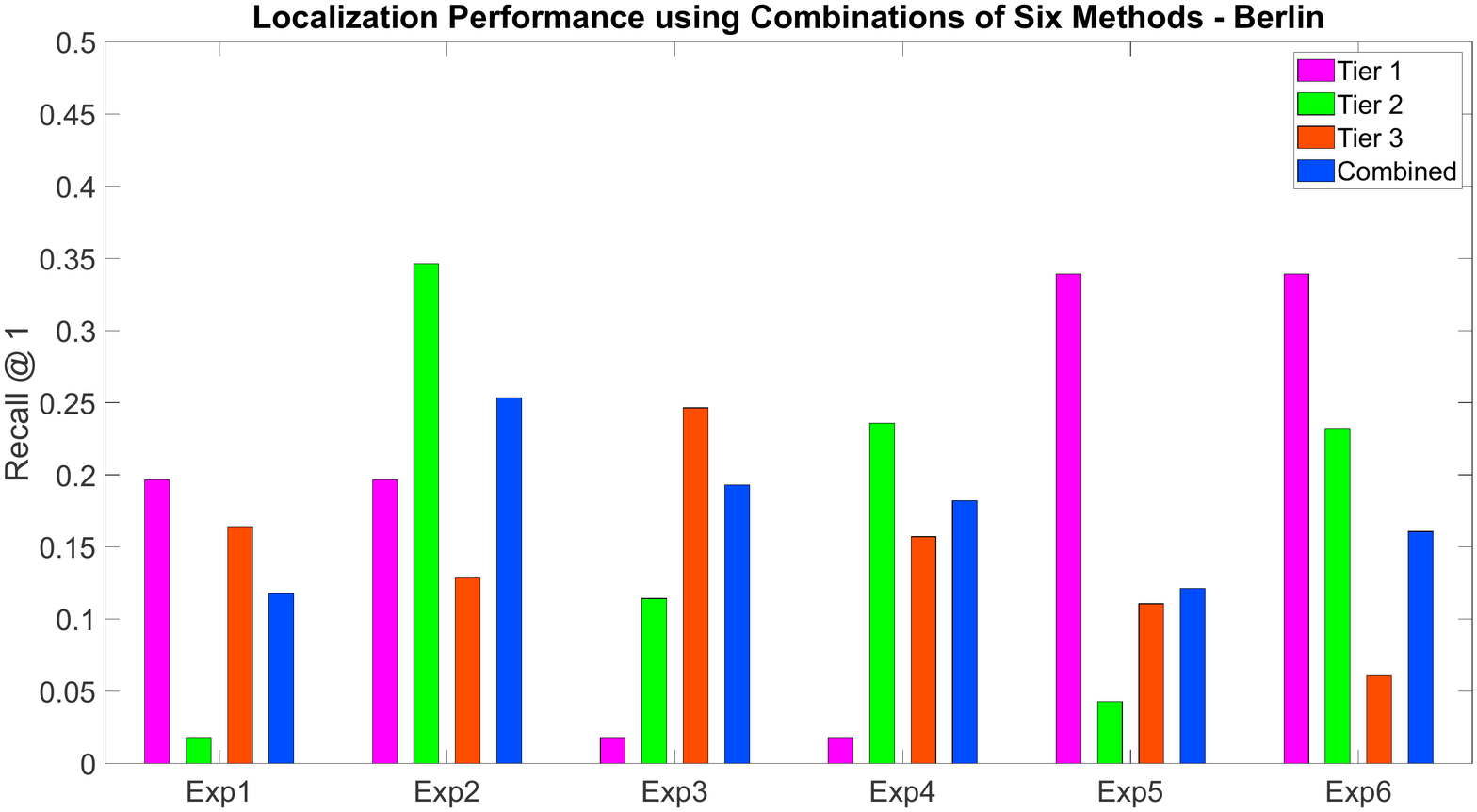}	%trim left,bottom,right,top  
\caption{Results for 6 methods on the Berlin train set. The Recall @ 1 is disjointed because of the failure of HOG, Hnet and Gist in this challenging environment.}
\label{SixMeth - Berlin}
\end{figure}

In Table \ref{SixTable}, we detail the hierarchy of methods used in each experimental combination for this section. We also provide the computation time to run each order of the six methods on the Nordland dataset, since some methods are more computationally intensive than others. In Figures \ref{SixMeth} and \ref{SixMeth - Berlin}, the Recall @ 1 for each combination is displayed. We found that HOG was particularly suited to localizing on Nordland, irrespective of the tiered position of the method. This is why in Experiments 1 and 2 the Recall @ 1 is higher in Tier 1 than Tier 3. Nonetheless, moving HOG and NetVLAD to a later tier still provides a localization improvement from 71.3\% in Tier 1 of Exp1 to 75.8\% in Tier 3 of Exp4. Using six methods on Berlin provides an interesting failure case scenario: the extremely poor performance of HybridNet, Gist and HOG often causes the ground-truth matching candidate to be rejected in the early tiers, even when returning the top 50 candidates from each method in the first tier. 

\begin{table}[h]
\caption{Six Method Combinations for Experiment}
\label{SixTable}
\centering
\resizebox{\linewidth}{!}{\begin{tabular}{|c|c|c|c|c|}
\hline\hline
\bfseries \thead{Experiment\\Number} & \bfseries Methods Tier 1 & \bfseries Methods Tier 2 & \bfseries Methods Tier 3 & \bfseries \thead{Compute Time\\per Frame (s)}\\
\hline
Exp1 & NetVLAD, HOG & Hnet, Gist & KAZE, OLO & 0.39\\
\hline
Exp2 & NetVLAD, HOG & KAZE, OLO & Hnet, Gist & 4.02\\
\hline
Exp3 & Hnet, Gist & NetVLAD, HOG & KAZE, OLO & 0.39\\
\hline
Exp4 & Hnet, Gist & KAZE, OLO & NetVLAD, HOG & 4.19\\
\hline
Exp5 & KAZE, OLO & Hnet, Gist & NetVLAD, HOG & 21.3\\
\hline
Exp6 & KAZE, OLO & NetVLAD, HOG & Hnet, Gist & 18.3\\
\hline
\end{tabular}}
\end{table}

\subsection{Study on Varying Number of Candidates Passed Between Tiers}

\begin{figure}[h]
\centering
\includegraphics[width=0.82\linewidth,trim=0cm 4.7cm 0cm 0cm,clip]{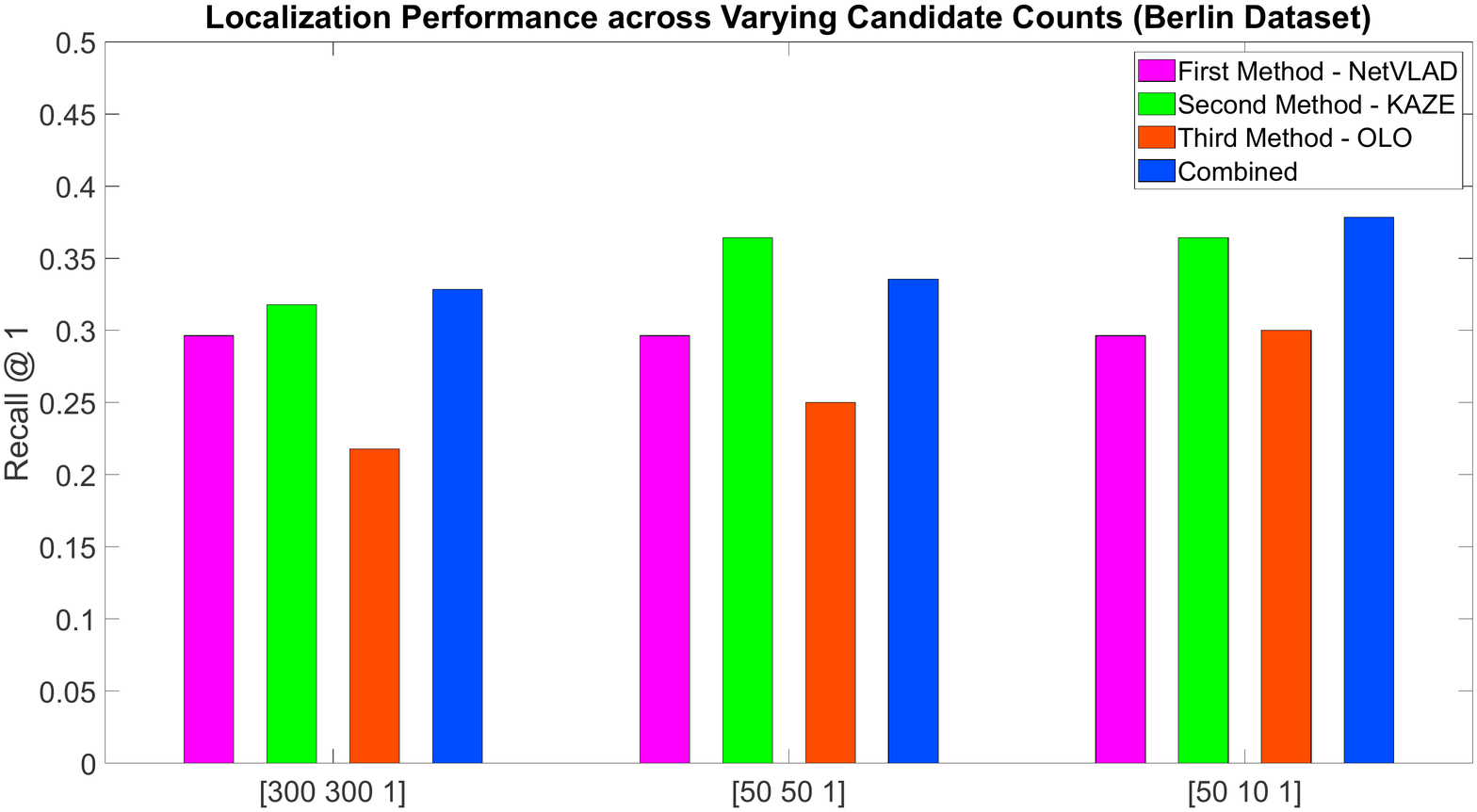}	%trim left,bottom,right,top  
\caption{Plot of Recall @ 1 across a varying candidate count, on the Berlin training set. We use the three methods NetVLAD, KAZE and OLO.}
\label{Var_Cands}
\end{figure}

We conclude our experiments on the training datasets by performing an investigation into a varying candidate count passed between hierarchies. The left-most bars in Figure \ref{Var_Cands} displays the localization performance for non-hierarchical multi-process fusion - each method in the hierarchy is given all the reference images. In the middle set of bars, the first method only passes the top 50 candidates to the second method. Finally the right-most bars show the results when the second method only passes the top 10 candidates to the third method. The improving performance demonstrates that hierarchical multi-process fusion is superior to simply fusing multiple methods in parallel.

\subsection{Evaluate Optimal Method Set on Test Sets}

We conclude our results by evaluating a set of selected methods on the two test datasets for Nordland and Berlin. We selected the set of methods with the highest Combined Recall @ 1 on the training sets. Therefore we selected Exp6 as the set of methods to evaluate on the Nordland test set, and the three method combination of NetVLAD, KAZE and OLO for Berlin. While we could have chosen the two method combination of NetVLAD and KAZE, which had equally good performance, the use of an additional method improves the robustness to dataset challenges introduced in the test set. 

\begin{figure}[h!]
\centering
\includegraphics[width=0.82\linewidth,trim=0cm 5.2cm 0cm 0cm,clip]{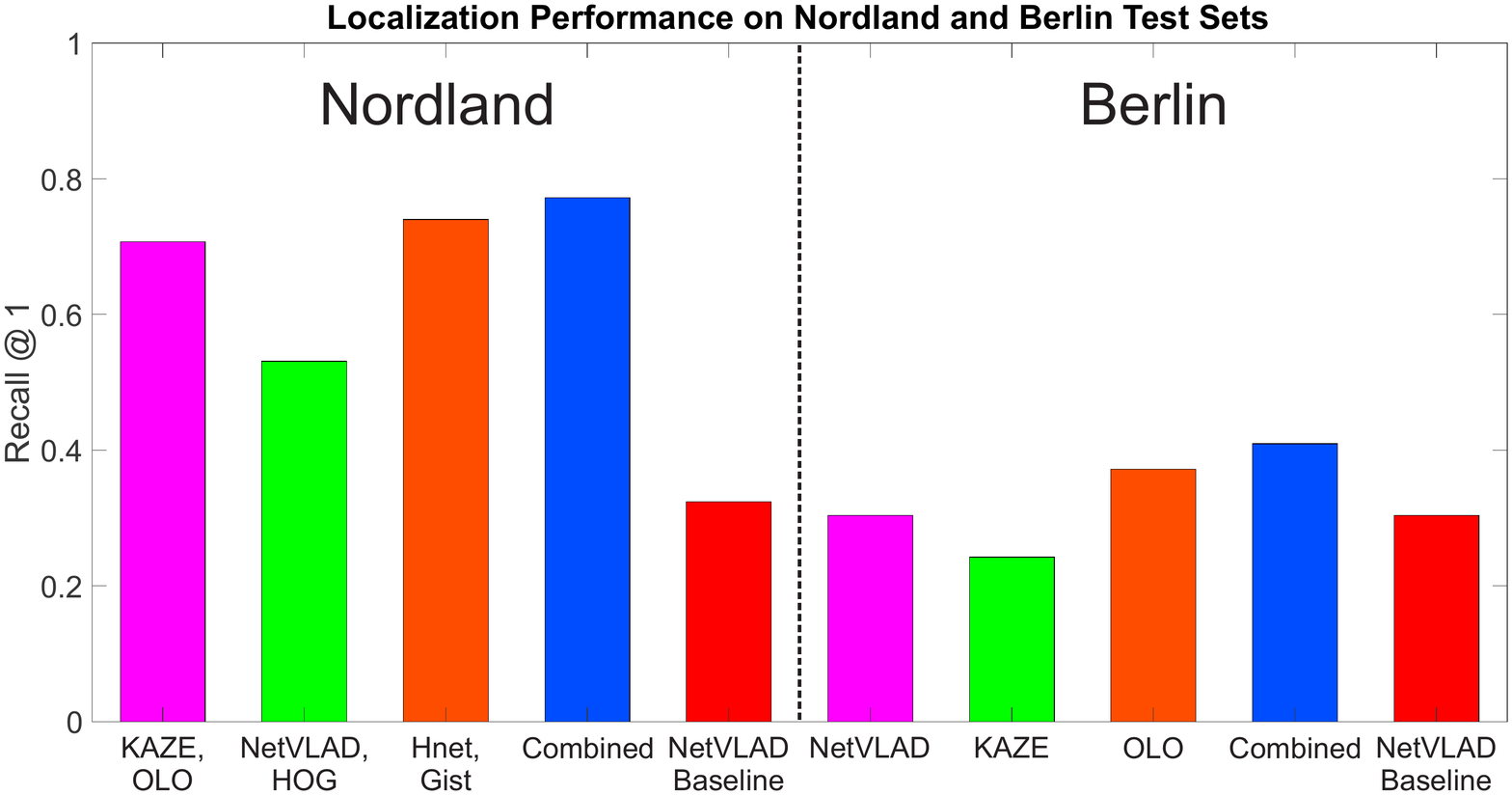}	%trim left,bottom,right,top  
\caption{In the five bars to the left, we evaluate our chosen set of six methods on the Nordland test set. In the right-hand bars, we evaluate the three best methods for Berlin using the test set.}
\label{EvalNord}
\end{figure}

In the Nordland test set (Figure \ref{EvalNord}), the combined algorithm produces the highest Recall @ 1 of 77.2\%. For the Berlin test set, the combined algorithm had a Recall @ 1 of 41.0\%. In both cases we improve upon the recall performance of the state-of-the-art algorithm NetVLAD. 

\section{Discussion and Conclusion}

In this paper we investigated the combination of multiple different image processing methods in a hierarchical structure, for the visual place recognition task. From our insights, we contribute a novel and high performing hierarchical framework for the localization task. Our results show that the combination of complementary methods in a hierarchy improves localization beyond any individual method, and the hierarchy, rather than a flat parallel structure, is key to this improvement. This can be observed by comparing the Recall @ 1 for a method in the first tier versus the second or third tiers in the hierarchy. We hypothesize that our approach works because each image processing method has its own varying criteria for which images are perceptually aliased with respect to the query image. By combining multiple methods in a hierarchy, an early tier method can filter out candidate images which would appear perceptually aliased to a later tier method.

By using a calibration training set we can remove or re-weight methods which perform poorly in the current environment. A profitable avenue of future work would be to add a source of weak ground-truth data or a second sensing modality, to decide on-the-fly whether a particular method needs to be omitted from the final place match decision process.

\clearpage

\bibliographystyle{IEEEtran}
\bibliography{MyCollection}

\end{document}